\documentclass{article}

\usepackage{PRIMEarxiv}

\usepackage[utf8]{inputenc} 
\usepackage[T1]{fontenc}    
\usepackage{hyperref}       
\usepackage{url}            
\usepackage{booktabs}       
\usepackage{amsfonts}       
\usepackage{nicefrac}       
\usepackage{microtype}      
\usepackage{lipsum}
\usepackage{fancyhdr}       
\usepackage{graphicx}       
\graphicspath{{media/}}     

\usepackage{url}
\usepackage{xcolor}

\usepackage{hyperref}

\usepackage{amsmath,amssymb,amsfonts}
\usepackage{algorithmic}
\usepackage{graphicx}
\usepackage{textcomp}

\usepackage{amsmath}
\usepackage{amssymb}
\usepackage{booktabs}
\usepackage{algorithm}
\usepackage{amsmath,pifont}
\usepackage{bm}
\usepackage{adjustbox}
\usepackage{capt-of}
\newcommand{\cmark}{\ding{51}}
\newcommand{\xmark}{\ding{55}}
\usepackage{caption}
\usepackage{subcaption}
\usepackage{multirow}

\usepackage[super]{nth}

\title{DrasCLR: A Self-supervised Framework of Learning Disease-related and Anatomy-specific Representation for 3D Medical Images
}

\author{
  Ke Yu$^*$ \\
  School of Computing and Information \\
  University of Pittsburgh \\
  \texttt{yu.ke@pitt.edu} \\
   \And
  Li Sun$^*$ \\
  Department of Electrical and Computer Engineering \\
  Boston University \\
  \texttt{lisun@bu.edu} \\
   \AND
  Junxiang Chen \\
  Department of Biomedical Informatics \\
  University of Pittsburgh \\
  \And
  Max Reynolds \\
  Department of Biomedical Informatics \\
  University of Pittsburgh \\
   \And
  Tigmanshu Chaudhary \\
  Department of Biomedical Informatics \\
  University of Pittsburgh \\
  \\
  \And
  Kayhan Batmanghelich \\
  Department of Electrical and Computer Engineering \\
  Boston University \\
}

\begin{document}
\maketitle
\def\thefootnote{*}\footnotetext{These authors contributed equally to this work}\def\thefootnote{\arabic{footnote}}

\begin{abstract}
\noindent Large-scale volumetric medical images with annotation are rare, costly, and time prohibitive to acquire. Self-supervised learning (SSL) offers a promising pre-training and feature extraction solution for many downstream tasks, as it only uses unlabeled data. Recently, SSL methods based on instance discrimination have gained popularity in the medical imaging domain. However, SSL pre-trained encoders may use many clues in the image to discriminate an instance that are not necessarily \emph{disease-related}. Moreover, pathological patterns are often subtle and heterogeneous, requiring the ability of the desired method to represent \emph{anatomy-specific} features that are sensitive to abnormal changes in different body parts. In this work, we present a novel SSL framework, named DrasCLR, for 3D medical imaging to overcome these challenges. We propose two domain-specific contrastive learning strategies: one aims to capture subtle disease patterns inside a local anatomical region, and the other aims to represent severe disease patterns that span larger regions. We formulate the encoder using conditional hyper-parameterized network, in which the parameters are dependant on the anatomical location, to extract anatomically sensitive features. Extensive experiments on large-scale computer tomography (CT) datasets of lung images show that our method improves the performance of many downstream prediction and segmentation tasks. The patient-level representation improves the performance of the patient survival prediction task. We show how our method can detect emphysema subtypes via dense prediction. We demonstrate that fine-tuning the pre-trained model can significantly reduce annotation efforts without sacrificing emphysema detection accuracy. Our ablation study highlights the importance of incorporating anatomical context into the SSL framework. 
\end{abstract}



\section{Introduction}
\label{sec:introduction}
While deep learning approaches have significantly advanced computer vision and many other fields~\cite{voulodimos2018deep, liu2020deep, pouyanfar2018survey}, efforts to apply these advancements to medical image analysis are still hampered by the scarcity of large-scale annotated datasets. Annotating medical images requires domain expertise and is a laborious and costly process, especially for 3D volumetric medical data. However, massive amounts of unlabeled raw images have been collected and stored in hospitals' picture archiving and communication systems (PACS) for decades. Recently, self-supervised learning (SSL) has become increasingly popular as a way to alleviate the annotation burden by exploiting the readily available unlabeled data~\cite{jing2020self, ohri2021review, you2022bootstrapping, you2022mine}. However, unlike supervised approaches, which use experts' annotations (\textit{e.g.}, disease labels, lesion segmentation masks) as supervision, self-supervised models are trained with limited supervision derived from the data itself, making it far more difficult to identify disease-related features from the data. Furthermore, certain lesions (\textit{e.g.}, early-stage tumors) may occupy only a small region in high-resolution volumetric medical images, and their visual patterns may vary depending on where they are located in the body. Thus, the desired self-supervised learning algorithm should be sensitive enough to capture local anatomical deformities. In this research, we propose DrasCLR: a novel framework for self-supervised learning of disease-related and anatomy-specific representation of 3D medical imaging. DrasCLR learns a patch-based dense representation that conditionally depends on the anatomical location of the center voxel. We extensively evaluate our method on chest computed tomography (CT) imaging because of its prominent role in the prevention, diagnostics and treatment of lung diseases. 

Self-supervised learning methods aim to provide useful feature representations for downstream tasks without human supervision, which is typically achieved by optimizing the model to solve a \emph{proxy} task. When designing a proxy task, the primary consideration is: \emph{what information in the data is important and what is not to the downstream tasks?} Early self-supervised approaches use heuristic-based pretext tasks to learn representations invariant to transformations that do not change the semantic meaning of the target labels~\cite{doersch2015unsupervised, zhang2016colorful, gidaris2018unsupervised}. More recent contrastive learning approaches~\cite{chen2020simple, he2020momentum} use \emph{instance discrimination task}, which consider every instance as a class of its own and train deep neural networks to discriminate pairs of similar inputs (augmented views of the same instance) from a selection of dissimilar pairs (different instances). In this setting, data augmentation guided by prior knowledge often plays a vital role in preserving task-relevant information~\cite{tian2020makes}. The sampling strategy for negative pairs is also crucial for the performance of contrastive learning methods. Recent studies~\cite{jin2018unsupervised, jeon2021mining} show that hard negative sampling guided by domain knowledge helps in preventing trivial solutions and improving the alignment of extracted features with human semantics.

Self-supervised representation learning of disease-related features in medical images is particularly challenging for two reasons. First, since disease-related features are often represented through subtle changes, an effective self-supervised learning method should be able to ignore large but irrelevant and non-informative information and focus on representing fine-grained features~\cite{holmberg2020self}. More specifically, this requires the encoder to be capable of representing small deviations from the appearance of normal anatomy. Second, because pathological tissues may only scatter in a few small regions, adequately representing local content is crucial for dense (voxel-level) prediction tasks such as anomaly detection and segmentation. Several self-supervised learning methods~\cite{zhou2019models, chaitanya2020contrastive, haghighi2021transferable} have been developed to learn local representations of 3D medical images. These methods use sub-volumes sampled from random locations in the image as inputs and train a single encoder with parameters shared across all locations. However, disease types and their visual patterns are often associated with anatomical locations. For example, pulmonary emphysema can be divided into three major subtypes (\textit{i.e.}, centrilobular, paraseptal, and panlobular) based on their visual characteristics and anatomical locations within the lung~\cite{smith2014pulmonary}. A more sophisticated framework for learning local representations should incorporate anatomical locations as prior information to account for the spatial heterogeneity of anatomical and pathological patterns.

\begin{figure*}
\centering
    \includegraphics[width = \textwidth]
    {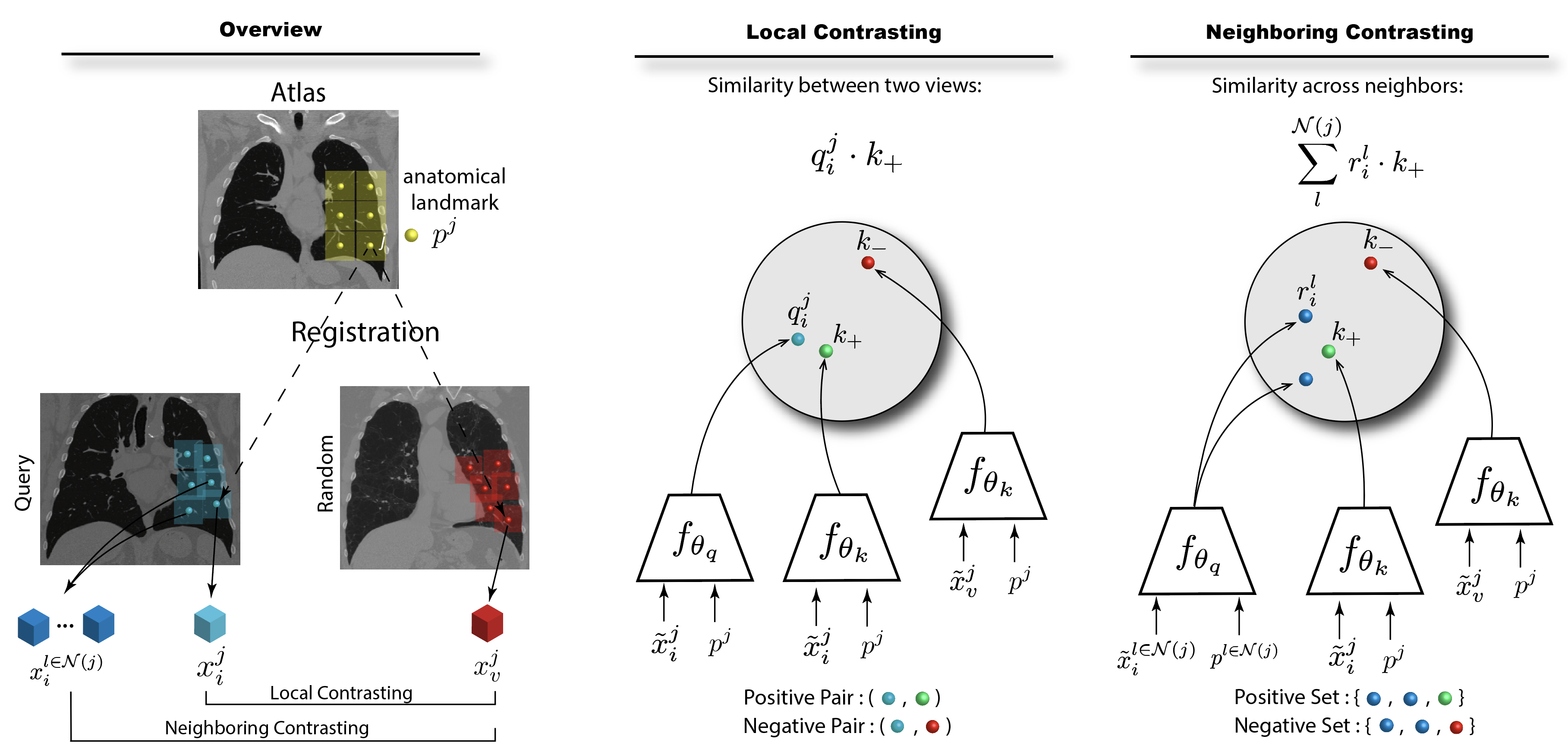}
    \caption{Schematic diagram of DrasCLR. \textbf{Left panel}: We represent a volumetric image with a collection of 3D patches registered with distinctive anatomical landmarks defined in an atlas image. We develop an encoder that generates location-specific representation using the patch and location of associated anatomical landmark as inputs. Our contrastive learning framework comprises two objectives. \textbf{Middle panel}: The first one aims to learn local representation from a single patch. \textbf{Right panel}: The second one aims to learn representations of larger patterns across neighboring patches. Both contrasting strategies incentivize the encoder to learn disease-related features by using patches of similar anatomy as \emph{hard} negative samples.
    }
    \label{fig:main}
\end{figure*}

In this research, we take inspiration from the aforementioned challenges and propose a novel contrastive learning framework for 3D medical imaging. In order to represent disease-related imaging features, we propose to combine two domain-specific contrasting strategies. The first strategy leverages the similarity across patients at the same anatomical location and aims to represent small disease patterns within a local (anchor) region. The second strategy takes advantage of anatomical similarities between the anchor and its nearby anatomical regions, with the goal of compensating for the first strategy by learning larger disease patterns that expand beyond the local region. We use small 3D patches to represent local anatomical regions. The effectiveness of both strategies depends on the difficulty of instance discrimination; as the anatomical similarity between the query and negative patches becomes greater, the encoder is forced to rely on subtle and disease-related features rather than normal anatomical features. To that end, we use image registration to obtain \emph{hard negative} patches from different subjects that are anatomically best aligned to the query patch. In particular, we obtain point-by-point correspondence between image pairs by mapping them to the same anatomical atlas. The coordinates in the atlas image can then be viewed as a standard set of anatomical locations. To incorporate anatomical locations into learned representations, we further develop a novel 3D convolutional layer whose kernels are conditionally parameterized through a routing function that takes the coordinates in atlas space as inputs. We call our unified framework \textbf{D}isease-\textbf{r}elated \textbf{a}natomy-\textbf{s}pecific \textbf{C}ontrastive \textbf{L}earning \textbf{R}epresentation (DrasCLR). The overview of our proposed approach is illustrated in Fig.~\ref{fig:main}. We conduct experiments on large-scale lung CT datasets. The results empirically show that our method outperforms baseline methods on both image-level and voxel-level tasks. Our code and pre-trained weights will be publicly available.

In summary, the major contributions of this paper are:
\begin{enumerate}
    \item We propose a novel framework for contrastive learning of disease-related representation for 3D medical images.
    \item We propose a novel 3D convolutional layer that encodes anatomical location-dependent information.
    \item We extensively validate our model on large-scale lung CT datasets and show that our method outperforms existing baselines for a wide range of image-level tasks. 
    \item We demonstrate the application of our method for voxel-wise emphysema detection and show that using our pre-trained model can significantly cut annotation costs without compromising detection accuracy.
\end{enumerate}

The paper is organized as follows: we first briefly introduce the related works in Section~\ref{sec:related_work}. In Section~\ref{sec:method}, we present the details of our proposed methodology. Implementation details and experimental results are described in Sections~\ref{sec:implementation} and ~\ref{sec:experiments}. Finally, we discuss the key findings and limitations of our work in Section~\ref{sec:discussions} followed by conclusions in Section~\ref{sec:conclusion}.

\section{Related Work}
\label{sec:related_work}
We review related works in four areas: (1) self-supervised learning approaches, including pretext task-based methods and contrastive learning methods, (2) applications of self-supervised learning in medical image analysis, (3) self-supervised learning methods that exploit anatomical context in medical images, and (4) conditionally parameterized networks.

\subsection{Self-supervised Learning}
Self-supervised learning has been shown to be an effective approach for learning semantically useful representations from large-scale unlabeled data without requiring human annotation~\cite{jing2020self, ohri2021review}. To generate supervisory signals from the data itself, a popular strategy is to present the model with various pretext tasks to solve. Commonly used pretext tasks include image inpainting~\cite{pathak2016context}, image colorization~\cite{zhang2016colorful}, relative position prediction~\cite{doersch2015unsupervised}, image jigsaw puzzle~\cite{noroozi2016unsupervised}, patch cut and paste~\cite{li2021cutpaste}, temporal order verification~\cite{misra2016shuffle}, geometric transformation recognition~\cite{gidaris2018unsupervised}, cross-modal correspondence~\cite{korbar2018cooperative, arandjelovic2017look}, and so on. These methods all have one thing in common: they build predictive-based pretext tasks using data's inherent structures, such as context similarity, spatial correlation, and temporal order. High-level semantic features are extracted in the process of accomplishing these tasks.

More recently, contrastive learning methods have emerged as one of the most popular self-supervised approaches due to their empirical success in computer vision~\cite{misra2020self, lee2021compressive}. The objective of contrastive training is to push learned representations to be similar for positive (similar) pairs and dissimilar for negative (dissimilar) pairs. This task is called \emph{instance discrimination}~\cite{wu2018unsupervised} and is often formulated using the InfoNCE loss~\cite{gutmann2010noise, van2018representation}. A variety of contrastive learning frameworks have been proposed, such as SimCLR~\cite{chen2020simple}, which uses the augmented view of the same input as positive samples and the augmented views of other samples in a minibatch as negative samples, and MoCo~\cite{he2020momentum}, which uses a slow moving average (\emph{momentum}) encoder and a dictionary that stores old negative representations to enable constructing very large batches of negative pairs. We extend MoCo as our contrastive learning paradigm. Rather than using a global dictionary, we develop a conditional memory bank that maintains distinct dictionaries of negative representations for each anatomical location. The design of sampling strategies for positive and negative pairs is a key driver to the success of contrastive learning~\cite{saunshi2019theoretical, tian2020contrastive}. Previous studies have demonstrated that encoders trained with harder negative pairs can represent more challenging features~\cite{jin2018unsupervised, jeon2021mining, robinson2020contrastive, kalantidis2020hard, robinson2021can}. We create negative pairs from examples with highly similar local anatomy to force the model to solve instance discrimination using more subtle visual features (\textit{e.g.} deviation from normal appealing tissues).

\subsection{SSL Applications in Medical Imaging}
Self-supervised learning is particularly useful for medical imaging analysis, in which labels are expensive to collect. Several studies have shown the effectiveness of self-supervised approaches in a variety of medical imaging analysis tasks such as disease diagnosis~\cite{shurrab2021self, li2021rotation, Azizi_2021_ICCV}, detection and localization~\cite{tajbakhsh2019surrogate, jiao2020self}, image segmentation~\cite{chaitanya2020contrastive, taleb20203d, ross2018exploiting, zhu2020rubik}, and image registration~\cite{li2018non}. Contrastive learning frameworks have been employed to leverage large-scale, unlabeled medical imaging data to produce pre-trained models. For example, ~\cite{sowrirajan2021moco} adopted MoCo as a pre-training approach to obtain high-quality representations for detecting diseases in chest X-rays; ~\cite{Azizi_2021_ICCV} extended SimCLR to train robust representations for dermatology condition classification and thoracic disease classification by using the availability of multiple views of the same pathology from the same patients. Our contrastive learning framework is built upon MoCo and is specifically designed to learn voxel-level representations that are sensitive to local anatomical deformities.

\subsection{Leveraging Anatomical Structure in SSL}
Medical images have consistent anatomy across patients, providing domain-specific cues for self-supervised representation learning. \cite{zhou2021models} introduced the Models Genesis, which learns image representation by recovering anatomical patterns from transformed sub-volumes extracted from CT images. \cite{haghighi2021transferable} extended the Models Genesis framework by adding a self-classification objective to enable the encoder to learn common anatomical semantics at similar body locations across patients. \cite{bai2019self} presented an anatomical position prediction task for learning segmentation features from cardiac magnetic resonance images. \cite{chaitanya2020contrastive} enhanced SimCLR by integrating two domain-specific contrasting strategies: incentivizing similar representations for volumetric image slices coming from similar anatomical areas, and incentivizing distinctive local representations for different anatomical regions coming from the same image. Our approach differs from the reviewed methods in two aspects: (1) We use image registration to improve the alignment of anatomical structures across patients, and (2) We leverage local anatomical similarity to create \emph{hard negative} samples to gain more information about tissue abnormalities.

\subsection{Conditionally Parameterized Networks}
In comparison to conventional neural networks with fixed weights, conditionally parameterized networks have weights computed as a function of the inputs. One such network is hypernetwork~\cite{ha2016hypernetworks}, which is typically a small network that outputs weights for a primary network. It has been used in functional image representation~\cite{klocek2019hypernetwork} and hyperparameter optimization~\cite{brock2017smash, lorraine2018stochastic, hoopes2021hypermorph}. Another prominent conditional parameterization technique is CondConv~\cite{yang2019condconv}, which creates convolutional kernels as a linear combination of experts with scalar weights dependent on input images. In our framework, the CondConv is modified to compute convolutional kernels as a function of the input anatomical locations. As a result, the learned representation at a given voxel depends on both its surrounding patch and its anatomical location.

\section{Method}
\label{sec:method}
We propose DrasCLR, a novel contrastive learning framework for 3D medical data. Our goal is to learn location-specific representations that are sensitive to tissue abnormalities. We start by aligning images to an \emph{anatomical atlas} using image registration and treating the image of each patient as a collection of 3D patches centered at a common set of anatomical locations. Our contrasting strategies are motivated by two domain-specific similarity cues: one leverages the similarity between patients at the same anatomical location, and the second leverages the similarity between nearby anatomical locations on the same image. In the following sections, we explain each component separately. The schematic diagram of the proposed method is shown in Fig.~\ref{fig:main}. The notations used in this paper are summarized in Table \ref{tbl:notation}.

\begin{table}[h]
 {
 \caption{Important notations in this paper.}
 \footnotesize
 \renewcommand{\arraystretch}{1.2}
 \centering
 \begin{tabular}{r   l}
     \toprule
     \multicolumn{2}{l}{
     \textbf{Models}} \\
     \midrule
     $e(\cdot, \cdot; \theta_1)$ & Image encoder. \\ 
     $g(\cdot; \theta_2)$ & MLP projection head.\\
     $f(\cdot, \cdot; \theta)$ & Network composed of $e$ and $g$, where $\theta = \{\theta_1, \theta_2\}$.  \\
     $r(\cdot)$ & Routing function used in Loc-CondConv.\\
     \midrule
     \multicolumn{2}{l}{\textbf{Functions}} \\
     \midrule
     $\phi_i(\cdot)$ & Transformation from the $i$-th image to the atlas space.\\
     $\phi^{-1}_i(\cdot)$ & Inverse transformation from atlas space to the $i$-th image.\\
     $\tilde t(\cdot)$ & Random augmentations.\\
     \midrule
     \multicolumn{2}{l}{\textbf{Variables}} \\
     \midrule
     $p^j$ & Location of the $j$-th anatomical landmark in the atlas space.\\
     $p_i^j$ & Location of the $j$-th anatomical landmark mapped in the $i$-th subject.\\
     $x_i^j$ & Patch of the $i$-th subject centering at the $j$-th anatomical landmark.\\
     $y_i^j$ & Representation of $j$-th patch in the $i$-th image used in downstream tasks.\\
     $y_i$ & Representation of the $i$-th image used in downstream tasks.\\
     $q_i^j$ & MoCo embedding of $j$-th  patch in the $i$-th query image.\\
     $k_+$ & MoCo embedding of the positive sample.\\
     $k_-$ & MoCo embedding of the negative sample.\\
     $r_i^l$ & MoCo embedding of $l$-th neighboring patch in the $i$-th image.\\
     ${\mathcal{N}(j)}$ & Neighboring patches of the $j$-th patch.\\
     $X_{Atlas}$ & The atlas image.\\
    \bottomrule
  \end{tabular} 

 \label{tbl:notation}
 }
 \end{table}

\subsection{Anatomical Alignment via Image Registration}
\label{sec:method_1}
We represent each volumetric image as a collection of 3D patches centered at a standard set of anatomical locations pre-defined on an anatomical atlas, with each patch corresponding to a distinct anatomical region of the lung. To align anatomical structures among patients, we first choose an image of a healthy subject to serve as the anatomical atlas, and then use image registration to obtain the subject-specific transformations that establish the point-by-point correspondence between the patients' images and the atlas image. Let $X_{\text{Atlas}}$ denote the atlas image, $x_i$ denote the image of patient $i$, the transformation $\phi_i$ is obtained by solving the optimization problem as follows:
\begin{equation}
\operatorname*{argmin}_{\phi_i} \hspace{0.15em} \texttt{Sim}\big(\phi_i(x_i), X_{\text{Atlas}}\big) + \texttt{Reg}(\phi_i),
\end{equation}
where $\texttt{Sim}(\cdot, \cdot)$ is mutual information similarity function and $\texttt{Reg}(\phi_i)$ is a regularization term to ensure the transformation is smooth. We perform the image registration using the Advanced Neuroimaging Tools (ANTs)~\cite{tustison2014large}.

After registration, we divide the lung region of the atlas image into $J$ evenly spaced three-dimensional patches with some overlap and define the patches' centers as the anatomical landmarks, denoted by $\{p^j\}_{j=1}^J$, where $j$ is the patch index and each $p^j \in \mathbb{R}^3$ is a coordinate in the \emph{Atlas space}. We apply the inverse transformation $\phi^{-1}_i$ to locate the anatomical landmarks on each patient's image and extract the corresponding patches for training. Formally, each patient's image $x_i$ is partitioned into a set of patches $\{x_i^j\}_{j=1}^J$ centered at $\{p_i^j\}_{j=1}^J$, respectively, where $x_i^j \in \mathbb{R}^{d \times d \times d}$, $p_i^j =\phi_i^{-1}(p^j)$ and $d$ is the dimension of patch. It is straightforward to show that patches with the same index across all patients correspond to the same anatomical region on the anatomical atlas:
\begin{equation}
\phi_i(p_i^j) = \phi_i\big(\phi_i^{-1}(p^j)\big)= p^j.
\end{equation}

\subsection{Conditionally Parameterized Convolutional  Layer}
\begin{figure}[t]
\centering
    \includegraphics[width = .35\textwidth]
    {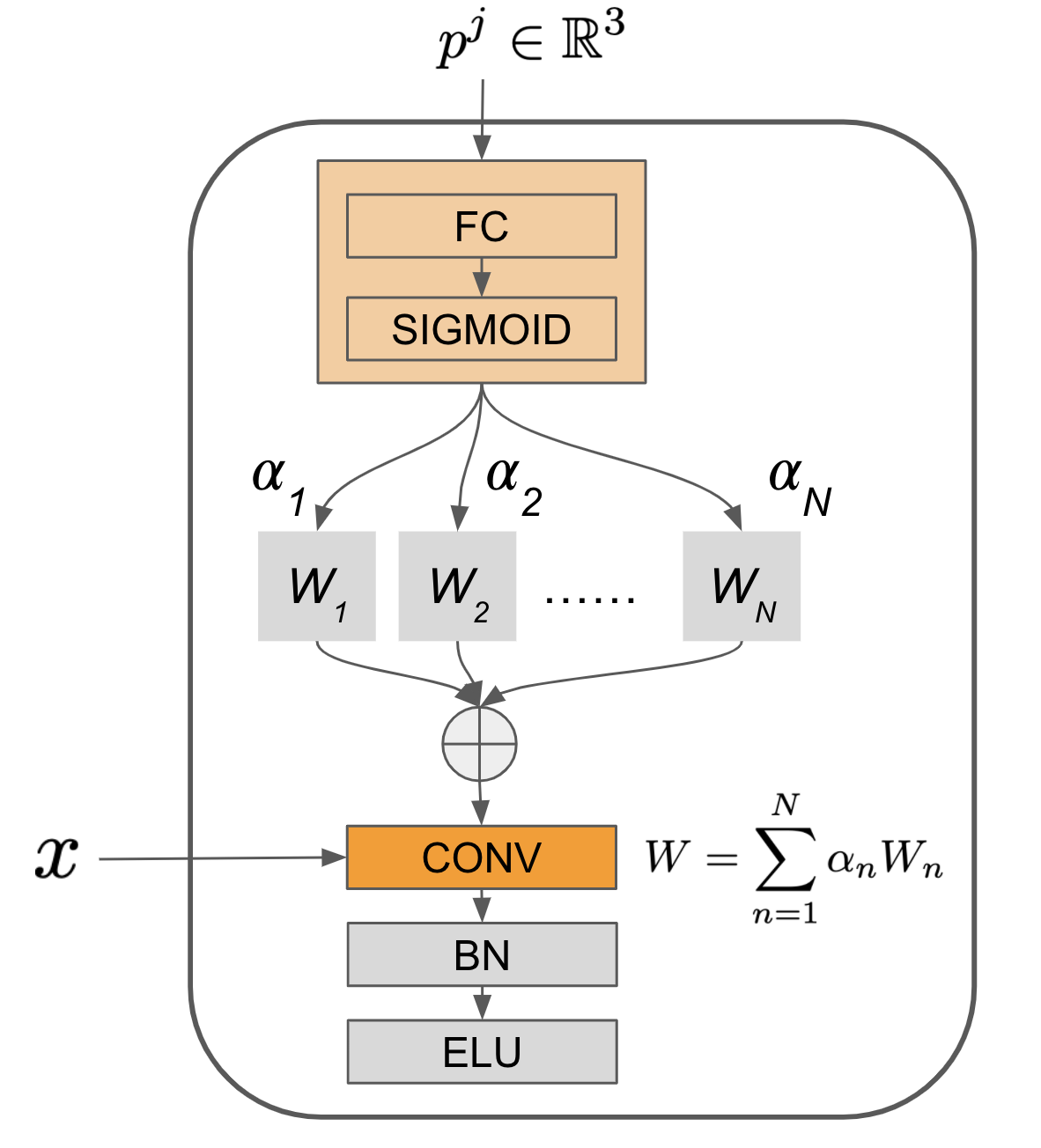}
    \caption{The architecture of the Loc-CondConv layer. The kernels $W$ are conditionally parameterized for each anatomical location $p^j$. The symbols $\alpha_n$ denote the routing weights. $x$ denotes the input from the previous layer.}
    \label{fig:loccondcov}
\end{figure}

Image patches from different anatomical locations have distinctive anatomical features and may be associated with different diseased tissue patterns. Standard convolutional  layers that apply the same kernels throughout the entire image may not be sufficient to accommodate spatial heterogeneity among patches at different locations. Inspired by CondConv~\cite{yang2019condconv}, we propose Loc-CondConv, a location-dependent, conditionally parameterized convolutional layer. Instead of using static convolutional kernels, we compute convolutional kernels as a function of the anatomical location. In particular, we parameterize the kernels in Loc-CondConv as a linear combination of $n$ convolutional kernels:
\begin{equation}
W = \alpha_1W_1 + \cdots + \alpha_N W_N,
\end{equation}
where $\{W_n\}_{n=1}^N$ are the same-sized convolutional kernels as in the regular convolutional layer and $\{\alpha_n\}_{n=1}^N$ are scalar weights computed via a routing function taking anatomical location as input. Specifically, we construct the routing function $r(
\cdot)$ using a fully-connected layer followed by a Sigmoid activation function:
\begin{equation}
r(p^j) = \sigma(p^j \times W_r),
\end{equation}
where $p^j$ is a coordinate in the \emph{Atlas space} and $W_r$ is a learnable weight matrix with dimension $3 \times N$, and $\sigma$ represents the sigmoid function. Fig.~\ref{fig:loccondcov} illustrates the architecture of Loc-CondConv. In the DrasCLR models, we replace all static convolutional layers with Loc-CondConv layers.

\subsection{Local Contrastive Loss}
\label{sec:local patch contrastive learning}
In contrastive learning, the model is trained to discriminate pairs of positive inputs from a selection of negative pairs. Recent studies show that selecting harder negative pairs is critical for the success of contrastive learning~\cite{saunshi2019theoretical, robinson2020contrastive}. The anatomical similarity between patients in the same lung region provides domain-specific cues for selecting hard negatives. More specifically, after registration alignment, any pair of patches centered at the same anatomical landmark, e.g. $x_i^j, x_v^j$ ($i  \neq v$), have highly similar local anatomy, forcing the encoder to discriminate them using more subtle visual features, such as pathological tissues, rather than shortcuts, such as the overall anatomical background or boundaries. 

With this motivation, we propose a \emph{local contrasting strategy}.  Formally, given a patch $x_i^j$, we generate two augmented views $\tilde x_i^j = \tilde t(x_i^j)$, where $\tilde t$ is random augmentations sampled from a set of transformations $\mathcal{T}$. These two augmented patches are considered as a positive pair. Each negative sample $\tilde x_v^j$ is generated as $\tilde x_v^j = \tilde t(x_v^j)$ by randomly sampling a patch in the same anatomical region $j$ from a different patient ($v \neq i$) and random augmentations $\tilde t \sim \mathcal{T}$. We adopt the MoCo~\cite{he2020momentum} as our contrastive learning paradigm. Specifically, we train two networks $f_{\theta_q}$, $f_{\theta_k}$ to map the positive pair $(\tilde x_i^j, \tilde x_i^j)$ and the negative pair $(\tilde x_i^j, \tilde x_v^j)$ to corresponding embeddings as follows:
\begin{equation}
\label{eq:5}
    q_i^j = f(\tilde{x}_i^j, p^j; \theta_q), \;
    k_+ = f(\tilde{x}_i^j, p^j; \theta_k), \;
    k_- = f(\tilde{x}_v^j, p^j; \theta_k),
\end{equation}
where $\theta_k = m\theta_k + (1-m)\theta_q$ and $m \in [0, 1)$ is a momentum coefficient. The network $f(\cdot, \cdot; \theta_q)$ is comprised of a feature extractor function $e(\cdot, \cdot; \theta_1)$, which accepts both patches and their corresponding anatomical landmarks as inputs, and a multilayer perceptron (MLP) projection head $g(\cdot; \theta_2)$, which maps the patch representations to the space where contrastive loss is applied. The equation can be written as $f(x_i^j, p^j; \theta_q)=g(e(x_i^j, p^j; \theta_1); \theta_2)$, where $\theta_q = \{\theta_1, \theta_2\}$. Finally, the \emph{local contrastive loss} per location is defined as:
\begin{equation}
\mathcal{L}_{l}^{j}=-\log\frac{\exp(q_i^j\cdot k_{+}/\tau)}{\exp(q_i^j\cdot k_{+}/\tau)+\sum^{{K}^-} \exp(q_i^j\cdot k_{-}/\tau)},
\end{equation}
where $K^-$ denotes the number of negative pairs and $\tau$ denotes the \emph{temperature} hyperparameter. 

\subsection{Neighboring Contrastive Loss}
The \emph{local contrastive loss} incentivizes representations to be sensitive to tissue abnormalities within local anatomical regions. Pathological tissues, however, may expand beyond the borders of a single patch. We develop a complementary contrasting strategy - \emph{neighboring contrasting} to allow the same encoder to learn disease patterns that may spread across multiple anatomical regions. For a given anatomical region $j$, we denote the indices of its $\ell$ nearest neighboring regions by $\mathcal{N}(j)$, its neighboring anatomical landmarks by $\{p^l\}_{l \in \mathcal{N}(j)}^{\ell}$, and the neighboring patches of $x_i^j$ on the same image by $\{x_i^l\}_{l \in \mathcal{N}(j)}^{\ell}$. The corresponding embeddings of the neighboring patches are given by:
\begin{equation}
    r_i^l = f(\tilde{x}_i^l, p^l; \theta_q), \; {l \in \mathcal{N}(j)},
\end{equation}
where $\tilde{x}_i^l = \tilde t (x_i^l)$ and $\tilde t \sim \mathcal{T}$. Instead of constructing positive and negative pairs, we construct positive and negative \emph{sets}, specifically,
\begin{align*}
& \text{positive set}: \left\{ \{\tilde x_i^l\}_{l \in \mathcal{N}(j)}^{\ell}, \: \tilde x_i^j\right\},  \\
& \text{negative set}: \left\{ \{\tilde x_i^l\}_{l \in \mathcal{N}(j)}^{\ell}, \: \tilde x_v^j\right\}, \: v \neq i.
\end{align*}
The \emph{neighboring contrastive loss} per location is define as:
\begin{equation}
\mathcal{L}_{n}^{j}=-\log\frac{\sum_l^{\mathcal{N}(j)}\exp(r_i^l\cdot k_{+}/\tau)}{\sum_l^{\mathcal{N}(j)}\exp(r_i^l\cdot k_{+}/\tau)+\sum_l^{\mathcal{N}(j)}\sum_{k_{-}}^{\mathcal{K}^-} \exp(r_i^l\cdot k_{-}/\tau)},
\end{equation}
where $k_+$ and $k_-$ are the same as defined in Eqn.~\ref{eq:5}. Minimizing this loss forces the encoder to extract similar visual features of the disease spreading across the patch $x_i^j$ and its neighboring patches $\{x_i^l\}_{l \in \mathcal{N}(j)}^{\ell}$. Additionally, by selecting random patches in the same anatomical region as the hard negatives, the encoder is prevented from using mismatched anatomy as a shortcut to perform this \emph{instance discrimination} task.

\subsection{Overall Model}
We train our model end-to-end by minimizing the combined \emph{local contrastive loss} and \emph{neighboring contrastive loss} and looping through each anatomical landmark. The overall loss function per location is defined as:
\begin{equation}
    \mathcal{L}^{j} = \mathcal{L}_{l}^{j} + \mathcal{L}_{n}^{j}.
\end{equation}

During inference time, the voxel-level representation can be obtained by:
\begin{equation}
    y_i^j = e(x_i^j, p^j; \theta_1), 
\end{equation}
where $e(\cdot, \cdot; \theta_1)$ is the trained encoder with Loc-CondConv layers. The image-level representation $y_i$ is obtained by averaging the representations of patches across all the anatomical landmarks. Formally, the representation at the image level is given by:
\begin{equation}
    y_i = \frac{1}{J} \sum_{j=1}^J e(x_i^j, p^j; \theta_1).
\end{equation}
Note that, at the time of inference, $p^j$ can be any point inside the atlas space and is not restricted to the predefined anatomical landmarks. In our experiments, we obtain image-level representations using only predetermined anatomical markers for computational efficiency.

\section{Implementation Details}
\label{sec:implementation}
We begin by extracting the lung regions from each CT scan using the lung segmentation method proposed by \cite{hofmanninger2020automatic}. We then choose the image of one healthy subject as the anatomical atlas and partition it into a grid of 3D patches with some overlap. This results in 581 patches, each with a size of $32 \times 32 \times 32$, that fully cover the lung in the atlas image. Anatomical landmarks are defined as the centers of these 581 patches on the atlas coordinate system. We use the image registration toolkit ANTs~\cite{tustison2014large} to obtain the forward and inverse affine transformations between each subject's (moving) image and the atlas (fixed) image.

We construct the encoder $e(\cdot, \cdot; \theta_1)$ using Loc-CondConv layers as the building blocks. Each Loc-CondConv layer contains $N$ 3D-convolutional kernels with size $3 \times 3 \times 3$ and is zero-padded on each side of the inputs by one pixel. We adopt batch normalization (BN)~\cite{ioffe2015batch} and ELU~\cite{clevert2015fast} activation following each Loc-CondConv. For the projection head $g(\cdot; \theta_2)$, we adopt a 2-layer MLP with ReLU activation. We set the number of nearest neighbors used in the neighboring contrastive loss as 2 based on an ablation study (Sec.~\ref{sec:ablation_neighbors}). We create data augmentations using  MONAI~\cite{MONAI_Consortium_MONAI_Medical_Open_2020}  package. The data augmentation includes random affine transforms (applied in the order of rotation, translation, and scale), Gaussian noise, and random image contrast adjustments. We optimize the networks using SGD with momentum = 0.9 and weight decay = $10^{-4}$. The learning rate is set to be $10^{-2}$ and is updated using a cosine schedule. We choose the batch size of 128. Following the practice in MoCo-v2~\cite{chen2020improved}, we set temperature $\tau$ to 0.2 and momentum coefficient to $0.999$. Unlike regular MoCo, which uses a single dictionary for negative samples, we develop a conditional memory bank that maintains separate dictionaries for anatomical landmarks, each of which has a size of 4096. For training, we select negative samples from the corresponding dictionary, which stores patch embeddings from the same anatomical location as the query patch. We perform self-supervised pretraining on the full dataset using four NVIDIA Tesla V100 GPUs, each with 32GB memory, for 48 hours or 20 epochs, whichever comes first. 

\section{Experiments}
\label{sec:experiments}
In this section, we take our DrasCLR pre-trained models and evaluate their performance in medical imaging tasks at both image and voxel levels. At the image level, we evaluate the effectiveness of the learned representation in disease phenotype prediction, disease severity classification, and survival analysis. At the voxel level, we first describe how our model can be used to produce voxel-wise segmentation masks. Using this approach, we then present the quantitative and qualitative results of subtype emphysema detection. Finally, we perform ablation studies to validate the importance of the proposed components in DrasCLR.

\subsection{Datasets}
We conduct the experiments on two large-scale datasets of 3D medical images, including the COPDGene dataset~\cite{regan2011genetic} and the MosMed dataset~\cite{morozov2020mosmeddata}. We apply the same data preprocessing procedure for images in both datasets. We begin by re-sampling all images into $1 mm^3$ isotropic resolution. We then threshold the Hounsfield Units (HU) to the intensity window of $[-1024,240]$ and normalize the intensity range to $[-1,1]$ by linear scaling. 

\subsubsection{COPDGene Dataset}
\label{sec:experiments_data_copd}
Chronic Obstructive Pulmonary Disease (COPD) is a chronic inflammatory lung disease that causes obstruction of lung airflow and is one of the leading causes of death worldwide. The COPDGene Study~\cite{regan2011genetic} is a multi-center observational study that collects imaging data, genetic biomarkers, and relevant phenotypes from a large cohort of subjects. In our study, we use a large set of 3D thorax CT images from 9,180 subjects for self-supervised pre-training. We use the spirometry measures, disease-related phenotypes, and survival status of the same cohort as the image-level labels in our experiments. On a subset of these CT scans, an experienced pulmonologist annotated the bounding boxes of subtypes of emphysema by clicking on locations surrounded by the pathological tissues~\cite{castaldi2013distinct,mendoza2012emphysema}. This procedure created 696 centrilobular emphysema bounding boxes from 153 subjects, and 243 paraseptal emphysema bounding boxes from 69 subjects. All these bounding boxes are of the same size ($32 mm^3$). We use this annotated subset to examine the performance of the DrasCLR pre-trained model for subtype emphysema detection.

\subsubsection{MosMed Dataset} 
The MosMed dataset contains 3D thorax CT images of 1,110 subjects from the municipal hospitals in Moscow, Russia~\cite{morozov2020mosmeddata}. 
Subjects in this dataset are classified into five grades (``Zero'', ``Mild'', ``Moderate'', ``Severe'', and ``Critical'') based on COVID-19 related CT findings and physiological measures, such as body temperature, respiration rate, blood oxygen saturation level (SpO2) and so on. Triage decisions are made based on the severity levels of the patients. For example, patients in the ``Moderate'' category only need to be followed up at home by a primary care physician, whereas patients in the ``Critical'' category are immediately transferred to the intensive care unit. We use the CT images in MosMed for model pre-training and use COVID-19 severity grades as classification labels in downstream analysis.

\subsection{Image Level Evaluation}
To assess how much disease-related information is preserved by the proposed method, we use the learned image-level representation to predict a wide range of clinical variables measured at the subject level, such as spirometry measurements, disease phenotypes, disease staging, and patients' survival rates.

\subsubsection{COPD Phenotype Prediction}
We begin by performing self-supervised pre-training with DrasCLR on the COPDGene dataset. Then, we use the learned image-level representations in downstream prediction tasks in a \emph{linear readout} fashion. In particular, we train linear regression models to predict two pulmonary function measures on the log scale, which are percent predicted values of Forced Expiratory Volume in one second (\texttt{FEV1pp}) and its ratio with Forced vital capacity (\texttt{$\text{FEV}_1 / \text{FVC}$}). We use $R^2$ scores as an evaluation metric for the regression analysis. In addition, we train multi-class logistic regression models to predict four categorical outcomes: (1) Global Initiative for Chronic Obstructive Lung Disease (GOLD) spirometric stage, a four-grade categorical variable indicating the severity of airflow limitation, (2) Centrilobular emphysema visual score (CLE), a six-grade categorical variable indicating the extent of emphysema in centrilobular, (3) Paraseptal emphysema visual score (Paraseptal), a three-grade categorical variable indicating the severity of paraseptal emphysema, and (4) Acute Exacerbation history (AE history), a binary variable indicating whether the patient has encountered at least one exacerbation event before enrolling in the study. For all classification tasks, we use accuracy as the evaluation metric. To account for human variability in annotation, for GOLD, CLE, and Paraseptal scores, we also report the proportion of times the predicted class fell within one class of the true score (denoted as \textit{1-off}).

We compare the performance of DrasCLR against both unsupervised and supervised approaches. The unsupervised baselines include: Models Genesis~\cite{zhou2021models}, MedicalNet~\cite{chen2019med3d}, MoCo (3D version on the entire volume)~\cite{he2020momentum}, Context SSL~\cite{sun2021context}, Divergence-based feature extractor~\cite{schabdach2017likelihood}, K-means algorithm applied to image features extracted from local lung regions~\cite{schabdach2017likelihood}, and Low Attenuation Area (LAA), a commonly used clinical descriptor. The supervised baselines include convolutional neural networks (CNN) that were separately trained to predict \texttt{FEV1pp}, GOLD and CLE scores using 2D slices as inputs (2D CNN)~\cite{gonzalez2018disease}, and Subject2Vec~\cite{singla2018subject2vec}, where a patch-based CNN model was first trained with \texttt{FEV1} and \texttt{$\text{FEV}_1 / \text{FVC}$} as joint supervised information, and the learned image representations were then used in other prediction tasks. We perform five-fold cross-validation for all experiments and report the average results along with standard deviations. Table~\ref{tbl:COPD} shows that the DrasCLR pre-trained model outperforms unsupervised baseline models in all metrics, with the exception of 1-off accuracy for Paraseptal emphysema, where the difference is within one standard deviation.
Our DrasCLR pre-trained model also outperforms the supervised baseline models, including Subject2Vec and 2D CNN, in terms of CLE, Paraseptal, and AE History predictions. For spirometry and COPD Staging, on which Subject2Vec were trained, the performance gap of our model is smaller compared to other unsupervised baseline models. 

Overall, these results suggest that image-level features extracted by the DrasCLR pre-trained model preserve richer information about COPD severity than other unsupervised baselines. When compared to supervised methods, our proposed method learns more generalizable features as it achieves higher predictive performance for a broader range of clinical variables, such as emphysema visual scores and AE history.

\begin{table*}[htp]
\caption{Results of phenotype prediction on the COPDGene dataset. We use R-Square for continuous measurements and accuracy for discrete scores. Results including the mean and standard deviation (mean$\pm$s.d.) are derived from 5-fold cross validation. Our DrasCLR model has the best or competitive performance on all phenotype prediction tasks when compared to seven unsupervised methods, and it generalizes better than the supervised method for predicting visual scores and AE history.}
 \begin{adjustbox}{max width=\textwidth}
 \centering
  \begin{tabular}{lc|cc|cc|cccc|c}
  \toprule
  \multirow{2}{*}{Method}&\multirow{2}{*}{Supervised}&\multicolumn{2}{|c|}{Spirometry}&\multicolumn{2}{|c|}{COPD Staging}&\multicolumn{4}{|c|}{Visual scores}& Acuity\\
    & &$\log$\texttt{FEV1pp}&$\log$\texttt{$\text{FEV}_1 / \text{FVC}$}&GOLD&GOLD \emph{1-off}&CLE&CLE \emph{1-off}&Paraseptal&Paraseptal \emph{1-off}&AE History\\
   \toprule
   Metric&  &\multicolumn{2}{|c|}{R-Square}& \multicolumn{7}{c}{\% Accuracy}  \\
   \midrule
 LAA-950 &\xmark&$0.44_{\pm.02}$&$0.60_{\pm.01}$&
 $55.8$&$75.7$&
 $32.9$&$77.7$&
 $33.3$&\bm{$87.6$}&$73.8$\\
 K-Means &\xmark&$0.55_{\pm.03}$&$0.68_{\pm.02}$&
 $57.3$&$82.3$&
 -&-&-&-&-\\
 Divergence-based &\xmark& $0.58_{\pm.03}$&$0.70_{\pm.02}$&
 $58.9$&$84.2$&
 -&-&-&-&-\\
 MedicalNet &\xmark&$0.47_{\pm.10}$&$0.59_{\pm.06}$&
 $57.0_{\pm1.3}$&$75.4_{\pm.9}$&
 $40.3_{\pm1.9}$&$69.6_{\pm1.6}$&$53.1_{\pm0.7}$&$81.8_{\pm0.8}$&
 $78.7_{\pm1.3}$\\
 ModelsGenesis&\xmark&$0.58_{\pm.01}$&$0.64_{\pm.01}$&
 $59.5_{\pm2.3}$&$82.9_{\pm1.3}$&
 $41.8_{\pm1.4}$&$77.0_{\pm1.5}$&
 $52.7_{\pm.5}$&$85.3_{\pm1.1}$&
 $77.8_{\pm.8}$\\
 MoCo &\xmark& $0.40_{\pm.02}$&$0.49_{\pm.02}$&
 $52.7_{\pm1.1}$&$67.6_{\pm1.4}$&
 $36.5_{\pm.7}$&$61.9_{\pm.9}$&$52.5_{\pm1.4}$&$79.7_{\pm1.2}$&
 $78.6_{\pm.9}$\\
 Context SSL&\xmark&$0.62_{\pm.01}$&$0.70_{\pm.01}$&
 $63.2_{\pm1.1}$&$83.6_{\pm.9}$&
 $50.4_{\pm1.3}$&$81.5_{\pm1.1}$&$56.2_{\pm1.1}$&$84.9_{\pm1.2}$&
 $78.8_{\pm1.3}$\\
 \midrule
2D CNN &\cmark&$0.53$&-&$51.1$&-&-&$60.4$&-&-\\
Subject2Vec &\cmark&\bm{$0.67_{\pm.03}$}&\bm{$0.74_{\pm.01}$}&
\bm{$65.4$}&\bm{$89.1$}&
$40.6$&$74.7$&$52.8$&$83.0$&
$76.9$\\
 \midrule
 Ours&\xmark&\underline{$0.63_{\pm.01}$}&\underline{$0.71_{\pm.01}$}&
 \underline{$65.0_{\pm.6}$}&\underline{$85.6_{\pm.6}$}&
 \underline{\bm{$53.9_{\pm.8}$}}&\underline{\bm{$86.3_{\pm.7}$}}&\underline{\bm{$58.4_{\pm.8}$}}&$87.0_{\pm.8}$&
 \underline{\bm{$78.9_{\pm1.3}$}}\\
  \bottomrule
  \\
   \multicolumn{11}{p{.5\textwidth}}{- indicates not reported.}\\
   \multicolumn{11}{p{\textwidth}}{Some baseline methods only report mean value without standard deviation in original manuscript.}\\
   \multicolumn{11}{p{\textwidth}}{The bold font is used to highlight the highest value for each column among \emph{all} methods.}\\
   \multicolumn{11}{p{\textwidth}}{The underline is used to highlight the highest value for each column among \emph{unsupervised} methods.}
  \end{tabular}
   \end{adjustbox}
   \label{tbl:COPD}
 \end{table*}
 
\subsubsection{Survival Analysis of COPD Patients}
We evaluate the effectiveness of DrasCLR in survival analysis for the COPDGene population.  We employ the Cox proportional hazards (CPH) model~\cite{cox1972regression} to predict patients' survival using the learned image-level representations while controlling for five potential confounders, including age, gender, race, smoking status, and packyear (calculated by multiplying the number of packs of cigarettes smoked per day by the number of years the person has smoked). We compare the performance of features extracted by our method against: (1) hand-designed imaging features, (2) imaging features retrieved by other machine learning methods, and (3) relevant clinical features, such as spirometry measures and the BODE index~\cite{celli2004body}. The hand-designed imaging features include CT metrics of emphysema, gas trapping, average wall thickness of hypothetical airway, and wall area percentage of segmental airways~\cite{martinez2006predictors}. All comparison baselines use the same CPH model and are controlled by including the same five confounding variables.

We report the results in terms of time-dependent concordance index ($C^{td}$), which estimates the model's risk ranking ability, at each of the censoring period quantiles. Table~\ref{tbl:concordance} shows that the survival model with our imaging features achieved concordance scores of 0.76, 0.75, and 0.74 at the \nth{25}, \nth{50}, and \nth{75} quantiles, respectively, outperforming baselines with imaging-only features retrieved by Models Gensis~\cite{zhou2021models}, Subject2Vec~\cite{singla2018subject2vec} and the hand-designed model. In comparison to clinical features, our method outperformed spirometry measures (0.76 vs 0.74) for risk stratification of near-term events before the \nth{25} quantile, slightly underperformed the BODE index (0.74 vs 0.75) at the \nth{75} quantile of censoring time, and achieved comparable accuracy otherwise. We also developed survival models with combined imaging and clinical features. The bottom rows of Table~\ref{tbl:concordance} show that the model using both our imaging features and BODE index achieved the highest concordance scores of 0.78, 0.77, and 0.77 at the \nth{25}, \nth{50}, and \nth{75} quantiles, respectively, demonstrating that the imaging representation learned by DrasCLR provides incremental predictive value for survival analysis of COPD patients.

\begin{table}[htp]
\caption{Time-dependent concordance index on the COPDGene dataset. Results are averages over five runs with bootstrapped standard errors. Our DrasCLR model performs the best when compared to other imaging representation approaches, and it provides incremental predictive value to clinical features.}
\vspace{10px}
\centering
\begin{adjustbox}{max width=0.7\textwidth}
\begin{tabular}{llccc}
\toprule
\multirow{2}{*}{\textbf{Feature}} & \multirow{2}{*}{\textbf{Method}} & \multicolumn{3}{c}{\textbf{Concordance Index}}\\ \cmidrule{3-5} 
 &  & $t = \nth{25}$ & $t = \nth{50}$ & $t = \nth{75}$\\ \midrule
\multirow{4}{*}{Imaging} & Hand-designed & $0.74_{\pm 0.02}$ & $0.73_{\pm 0.01}$ & $0.74_{\pm 0.01}$ \\
 & Models Genesis & $0.72_{\pm 0.02}$ & $0.7_{\pm 0.01}$ & $0.72_{\pm 0.01}$\\
 & Subject2Vec & $0.72_{\pm 0.02}$ & $0.72_{\pm 0.01}$ & $0.72_{\pm 0.03}$\\
 & Ours & $0.76_{\pm 0.02}$ & $0.75_{\pm 0.01}$ & $0.74_{\pm 0.01}$\\ \midrule
\multirow{2}{*}{Clinical} & Spirometry & $0.74_{\pm 0.02}$ & $0.75_{\pm 0.01}$ & $0.74_{\pm 0.01}$\\
 & BODE & $0.76_{\pm 0.01}$ & $0.75_{\pm 0.01}$ & $0.75_{\pm 0.00}$\\ \midrule
\multirow{3}{*}{Imaging + Clinical} & Hand-designed + BODE & $0.76_{\pm 0.02}$ & $0.76_{\pm 0.01}$ & $0.76_{\pm 0.01}$\\
 & Ours + Spirometry & $0.77_{\pm 0.02}$ & $0.76_{\pm 0.00}$ & $0.76_{\pm 0.01}$\\
 & Ours + BODE & $\bm{0.78_{\pm 0.01}}$ & $\bm{0.77_{\pm 0.01}}$ & $\bm{0.77_{\pm 0.00}}$\\ \bottomrule
\end{tabular}
\label{tbl:concordance}
\end{adjustbox}
\end{table}

\subsubsection{COVID-19 Severity Prediction}
We first pre-train a model with DrasCLR on the CT scans in the MosMed dataset. Then, we freeze the encoder and train a linear classifier to predict the severity of COVID-19, a categorical variable with five grades. The DrasCLR pre-trained model is compared to unsupervised baseline models, including MedicalNet~\cite{chen2019med3d}, Models Genesis~\cite{zhou2021models}, MoCo~\cite{he2020momentum}, Context SSL~\cite{sun2021context}, and one supervised CNN model that uses entire 3D images as inputs (3D CNN). To evaluate classification performance, we perform five-fold cross-validation and report the average test accuracy along with standard deviations.
 
 \begin{table}[t]
 \caption{Classification of 5-grade COVID-19 severity on the MosMed dataset. The results are the means and standard deviations of accuracy for 5-fold cross validation. Our DrasCLR model leads the best performance over both unsupervised and supervised approaches.}
 \vspace{10px}
 \centering
  \begin{tabular}{lcc}
  \toprule
   Method&Supervised&\% Accuracy\\
   \midrule
  3D CNN &\cmark&$61.2_{\pm3.5}$\\
  MedicalNet &\xmark&$62.1_{\pm3.3}$\\
  Models Genesis &\xmark&$62.0_{\pm3.5}$\\
  MoCo &\xmark&$62.1_{\pm3.3}$\\
  Context SSL&\xmark&$65.3_{\pm3.2}$\\
 \midrule
Ours w/o Neighbor Contrast&\xmark&$62.6_{\pm2.4}$\\
Ours&\xmark&\bm{$65.4_{\pm2.5}$}\\
  \bottomrule
  \end{tabular}
  \label{tbl:RU}
 \end{table}

Table~\ref{tbl:RU} shows that the DrasCLR pre-trained model outperforms the unsupervised baseline models with the exception of ContextSSL, whose average test accuracy is close to ours (65.3 vs. 65.4). Both ContextSSL and our methods leverage the context between neighboring anatomical regions for representation learning. ContextSSL incorporates this information via a graph neural network, whereas our method uses a neighboring contrasting strategy. The ablation study results in Table~\ref{tbl:RU}'s bottom rows show that leveraging anatomical context from large regions is useful for categorizing COVID-19 severity. With the neighboring contrastive loss, the COVID-19 severity prediction accuracy increases by $2.8\%$. Interestingly, we found that the supervised 3D CNN model performs the worst, suggesting that directly extracting features from the entire volume may have resulted in the loss of fine-grained information at local anatomy. It is also possible that the supervised model may not converge properly or becomes overfitted due to the small amount of training data.

\subsection{Voxel Level Evaluation} 
To show that the DrasCLR pre-trained model encodes fine-grained information at local anatomy, we demonstrate its ability to detect two subtypes of emphysema (\textit{i.e.}, centrilobular and paraseptal emphysema), which are prevalent in different pulmonary regions. We perform the experiments in three aspects. First, we conduct a quantitative evaluation for emphysema detection via voxel-wise classification. Second, we show the qualitative results of predicted emphysema masks for COPD patients at different stages. Third, we perform emphysema detection on a group of randomly selected subjects and show the relationship between the detected emphysema volume and the patient's COPD stage. In addition, we demonstrate that our method can reduce annotation efforts through transfer learning. 

\subsubsection{Emphysema Detection via Dense Classification}
\label{sec: emphysema_detection}
We propose to use dense or voxel-level classification for emphysema detection. In particular, we first pre-train a model with DrasCLR on all CT scans from the COPDGene dataset. We then fine-tune the model in a binary classification task to discriminate between emphysema-annotated patches and healthy patches. The healthy patches are sampled from random subjects with the criteria that the subjects' GOLD scores are equal to zero, and no centrilobular or paraseptal emphysema is found based on their image-level visual scores. We have evaluated two fine-tuning schemes: (1) employing the pre-trained encoder as a fixed feature extractor and training a new linear classifier (named as \textit{linear readout}), and (2) appending a linear classifier to the pre-trained encoder and fine-tuning all layers in the network (named as \textit{full fine-tuning}). During inference time, the fine-tuned model is used to perform voxel-wise classification of emphysema. For a given patch, the detection is positive if $25\%$ voxels within it have a predicted probability of emphysema greater than 0.5; otherwise, the detection is negative.

Table~\ref{tbl:recall} presents the quantitative performance of our method and selected baselines for emphysema detection. Under the linear readout scheme, we compare the performance of our model to logistic regression models using patch-level LAA features and learned features from Models Genesis~\cite{zhou2021models}. Under the full fine-tuning scheme, we compare our model's performance to a patch-based CNN that is trained on the same set of patches from scratch. 
The results in Table~\ref{tbl:recall} show that, compared to unsupervised baselines, our method achieves superior F1 scores for both centrilobular and paraseptal emphysema detection under linear readout. Through full fine-tuning, our method achieves the best emphysema detection performance across all evaluation metrics.  
 
\begin{table*}[t]
\caption{Evaluation for subtype emphysema detection. Results are the means and standard deviations (mean$\pm$s.d.) of F1, precision and recall scores for 5-fold cross validation. With linear readout, our DrasCLR outperforms unsupervised baselines. With full fine-funing, our DrasCLR outperforms the supervised baseline.}
\vspace{5px}
\centering
\begin{tabular}{lc|ccc|ccc}
\toprule
&& \multicolumn{3}{c|}{Centrilobular}
&\multicolumn{3}{c}{Paraseptal}\\
\midrule
&Supervised&F1&Precision& Recall&F1& Precision& Recall\\
\midrule
Patch LAA&\xmark&$0.77_{\pm.03}$&\underline{$0.94_{\pm.02}$}&$0.65_{\pm.04}$&
      $0.70_{\pm.06}$&\underline{$0.83_{\pm.06}$}&$0.61_{\pm.09}$\\
Models Genesis&\xmark&$0.71_{\pm.01}$&$0.58_{\pm.02}$&\underline{$0.93_{\pm.03}$}&
           $0.67_{\pm.03}$&$0.64_{\pm.16}$&$0.82_{\pm.23}$\\
Ours w/ Linear Readout &\xmark&\underline{$0.82_{\pm.03}$}&$0.75_{\pm.04}$&$0.91_{\pm.02}$&
 \underline{$0.82_{\pm.03}$}&$0.73_{\pm.05}$&\underline{$0.95_{\pm.01}$}\\
\midrule
Patch-based CNN&\cmark&$0.91_{\pm.03}$&$0.92_{\pm.01}$&$0.89_{\pm.05}$&
            $0.82_{\pm.04}$&$0.77_{\pm.05}$&$0.88_{\pm.02}$\\
Ours w/ Full Fine-tuning &\cmark&\bm{$0.98_{\pm.01}$}&\bm{$0.97_{\pm.01}$}&\bm{$0.99_{\pm.01}$}&
            \bm{$0.99_{\pm.01}$}&\bm{$0.99_{\pm.02}$}&\bm{$1.00_{\pm.00}$}\\
\bottomrule

\multicolumn{8}{p{0.95\textwidth}}{The bold font is used to highlight the highest value for each column among \emph{all} methods.}\\
\multicolumn{8}{p{0.95\textwidth}}{The underline is used to highlight the highest value for each column among \emph{unsupervised} methods.}\\
\end{tabular}
\label{tbl:recall}
\end{table*}
 
\begin{figure*}
\centering
    \includegraphics[width = 0.9\textwidth]
    {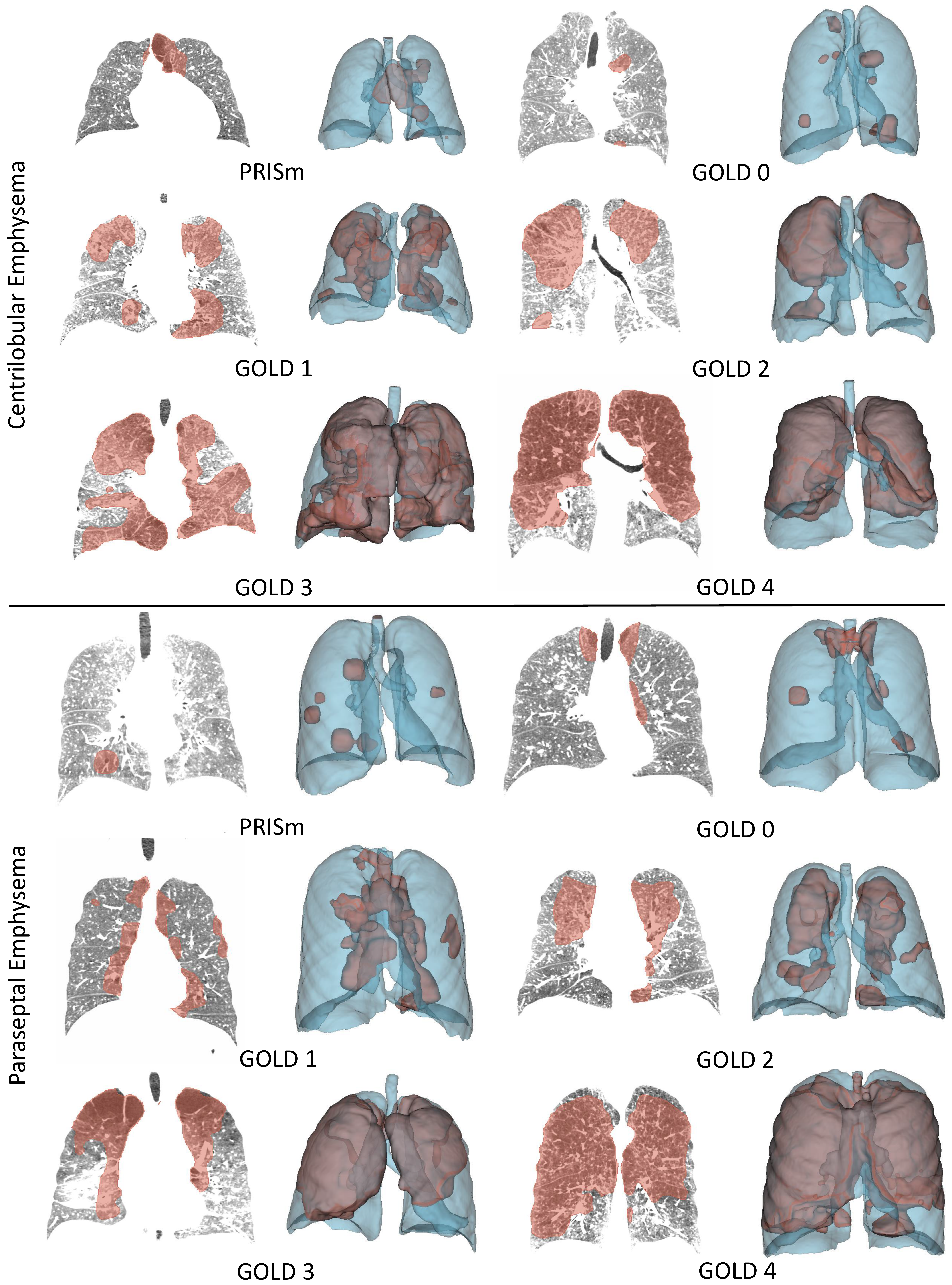}
    \caption{Examples of predicted dense emphysema binary masks for subjects with different GOLD scores. The top three rows show the predicted regions of centrilobular emphysema, and the bottom three rows show the predicted regions of paraseptal emphysema. The intensity range is set as [-1060, -825] to better illustrate the emphysema. The predicted emphysema regions are plotted in red, and the lung regions are plotted in blue. As the severity of COPD increases (higher GOLD score), the detected region increases in both subtypes of emphysema. In addition, the predicted emphysema regions correspond to the clinical description of their subtypes.
    }
    \label{fig:emph_volume}
\end{figure*}

To qualitatively demonstrate the outcomes of our method, we create voxel-wise emphysema segmentation for subjects at different stages of COPD. In particular, we first use the full fine-tuning model to estimate the probability of emphysema in a sliding-window fashion with a step size of 1 voxel. We then use a 0.5 threshold to map voxels with emphysema probability greater than or equal to the threshold to 1 and all other voxels to 0. Fig.~\ref{fig:emph_volume} shows the predicted segmentation masks of two subtypes of emphysema of varying COPD stage in coronal and 3D views. We find that as the COPD severity increase (higher GOLD score), the volume of detected emphysema region increases in both subtypes. Furthermore, segmentation masks of GOLD scores 1 and 2 show a clear heterogeneity in the regional distribution of emphysema in the lung between these two subtypes. The regions of predicted segmentation are consistent with the clinical definition of emphysema subtypes, where centrilobular emphysema is commonly described as an abnormal enlargement of airspaces centered on the respiratory bronchiole~\cite{leopold1957centrilobular} and paraseptal emphysema refers to emphysematous change adjacent to a pleural surface~\cite{heard1979morphology}.

\begin{figure*}[h]
\centering
\begin{subfigure}{0.48\textwidth}
    \centering
    \includegraphics[width = 1.\textwidth]
    {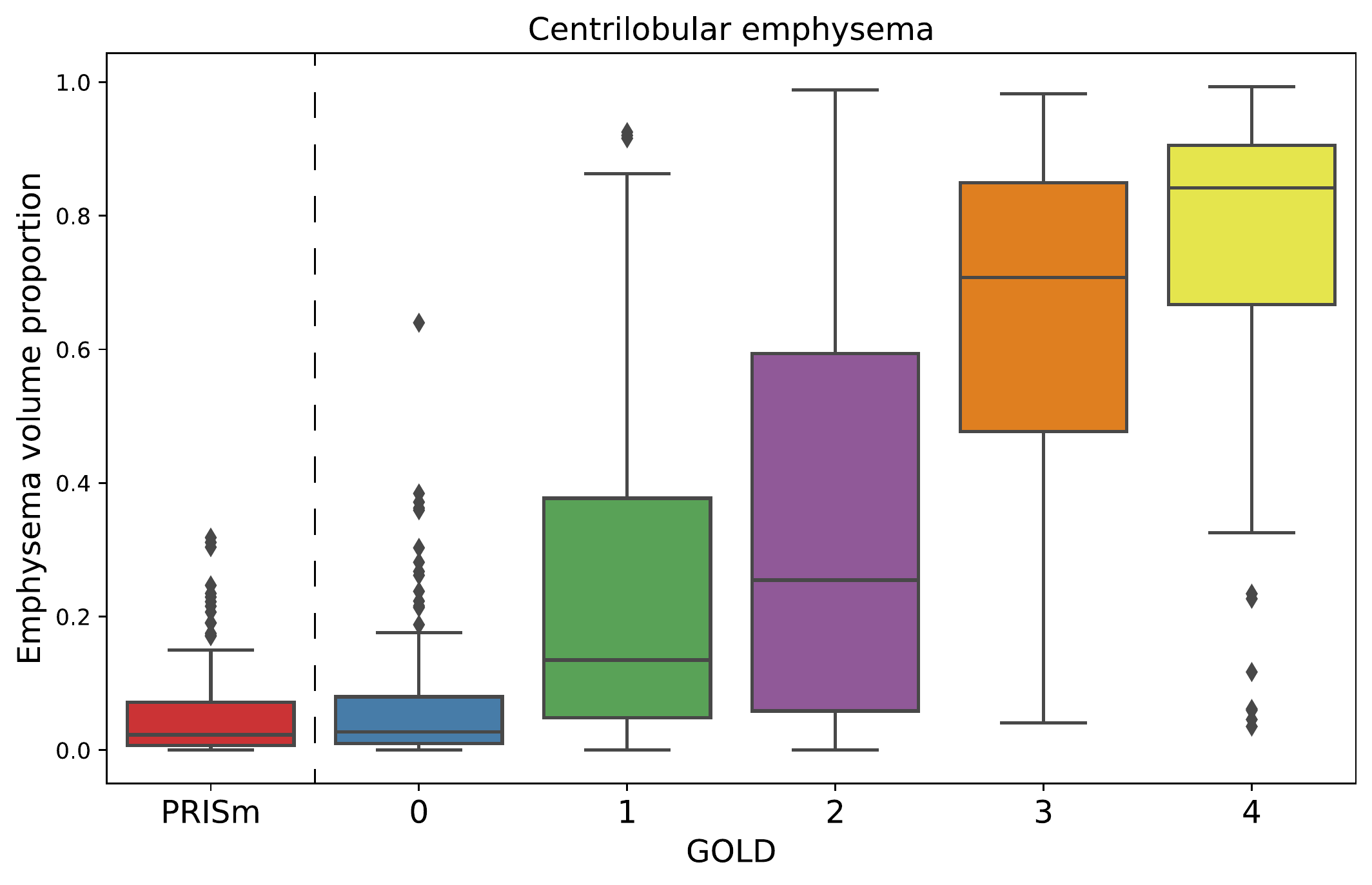}
\end{subfigure}
\begin{subfigure}{0.48\textwidth}
    \centering
    \includegraphics[width = 1.\textwidth]
    {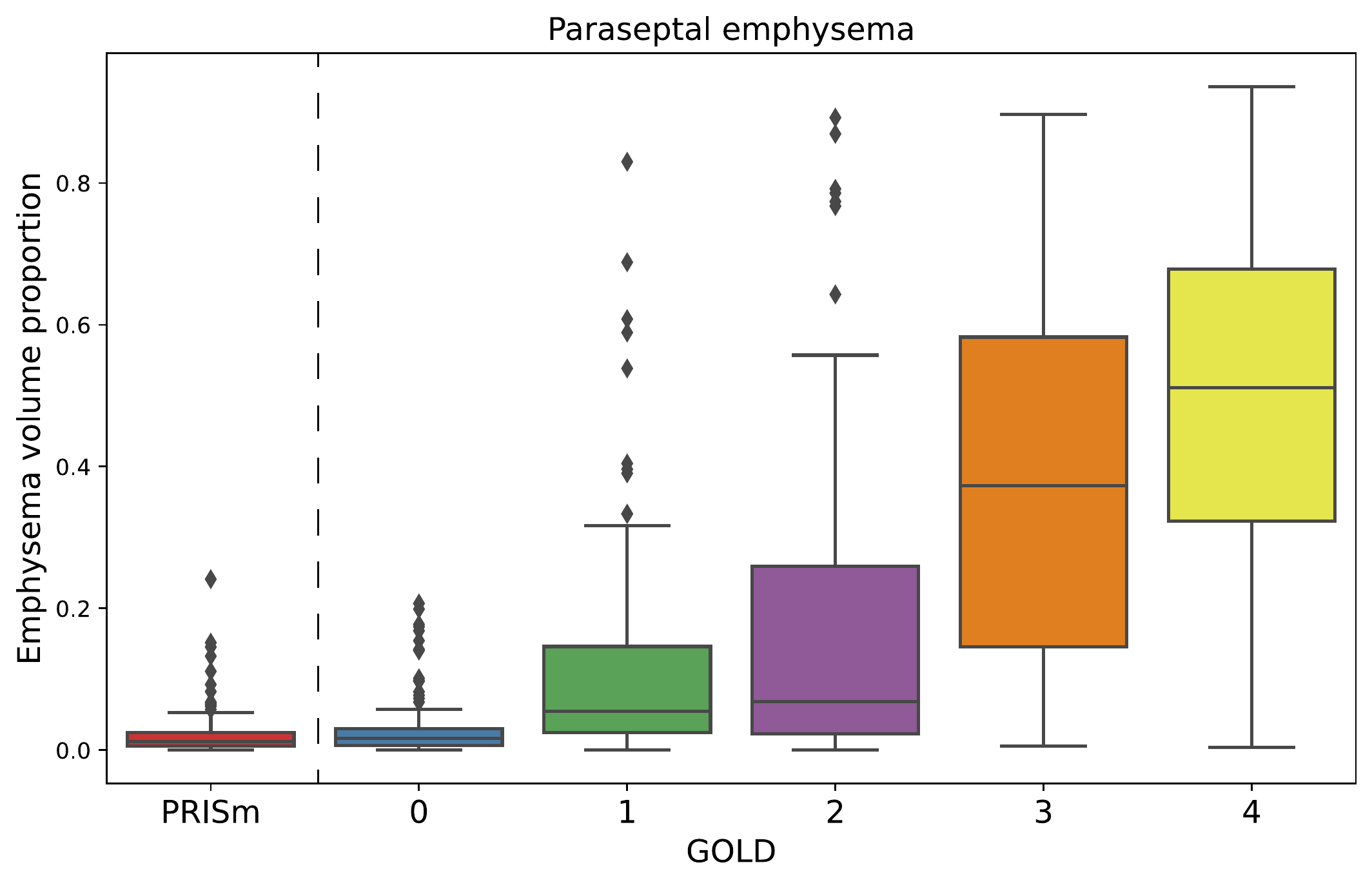}
\end{subfigure}
    \caption{Comparison of predicted volume proportion of centrilobular (left) and paraseptal (right) emphysema for subjects with different GOLD scores. A higher GOLD score indicates a more severe stage of COPD. The results show that with an increasing GOLD score, the predicted emphysema volume proportion in CT scan becomes higher.
    }
    \label{fig:volume_proportion}
\end{figure*}

Finally, we analyze the correlation between the total emphysema detected in 3D images and the subjects' COPD stages. In particular, we randomly select 500 subjects from the COPDGene dataset and use the fully fine-tuned model to make voxel-wise emphysema classification on their CT scans. Then, we aggregate all voxels in a CT scan to determine the fraction of voxels with a predicted probability of emphysema greater than 0.5. The box plots in Fig.~\ref{fig:volume_proportion} represent the distributions of detected emphysema proportion against GOLD scores, as well as the group with preserved ratio impaired spirometry (PRISm). Subplots of both centrilobular and paraseptal emphysema show positive correlations between the detected emphysema and patient's COPD severity. 

\subsubsection{Improve Annotation Efficiency via Transfer Learning}
\begin{figure*}[h]
\centering
\begin{subfigure}{0.48\textwidth}
    \centering
    \includegraphics[width = 1.\textwidth]
    {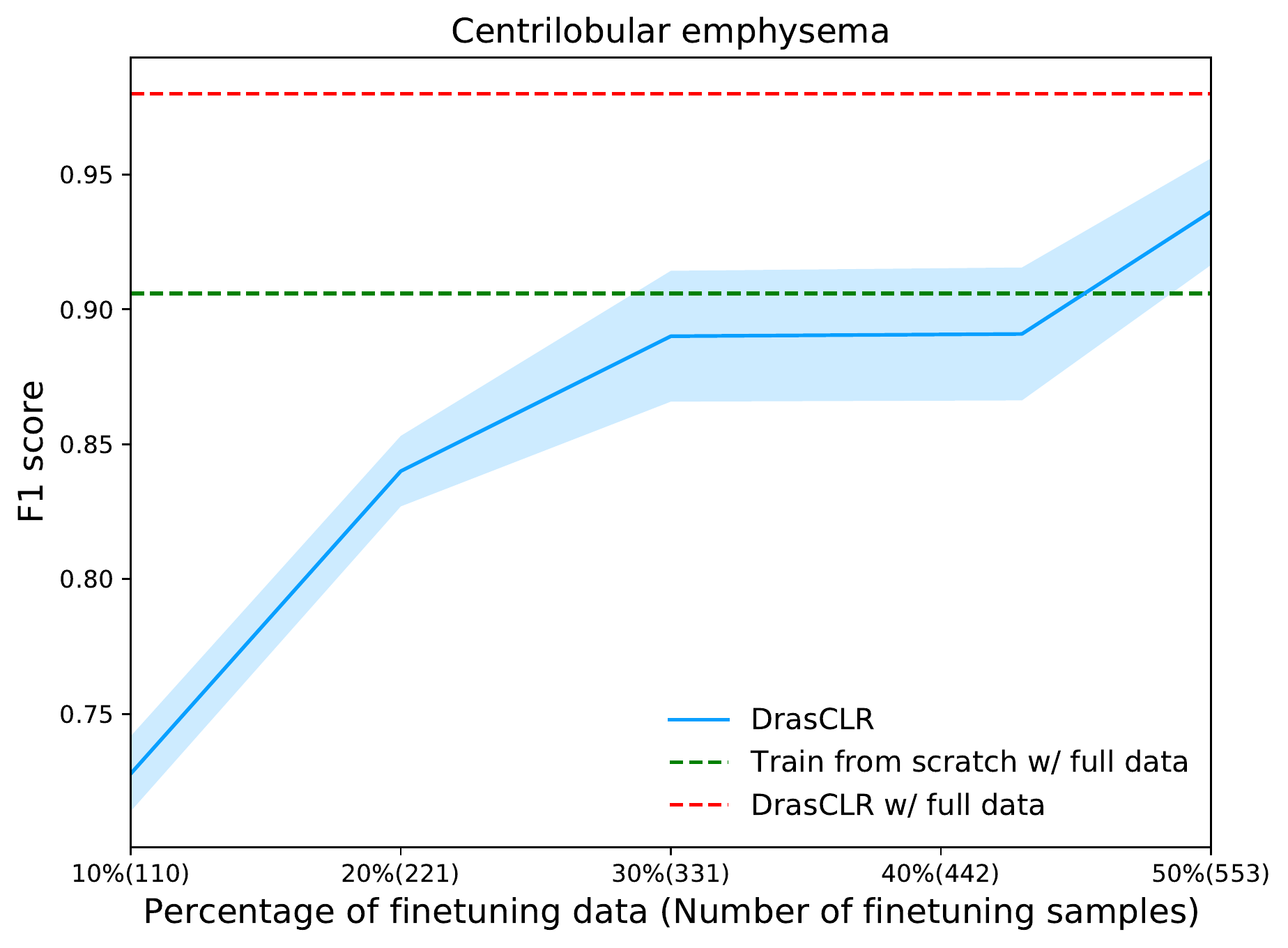}
\end{subfigure}
\begin{subfigure}{0.48\textwidth}
    \centering
    \includegraphics[width = 1.\textwidth]
    {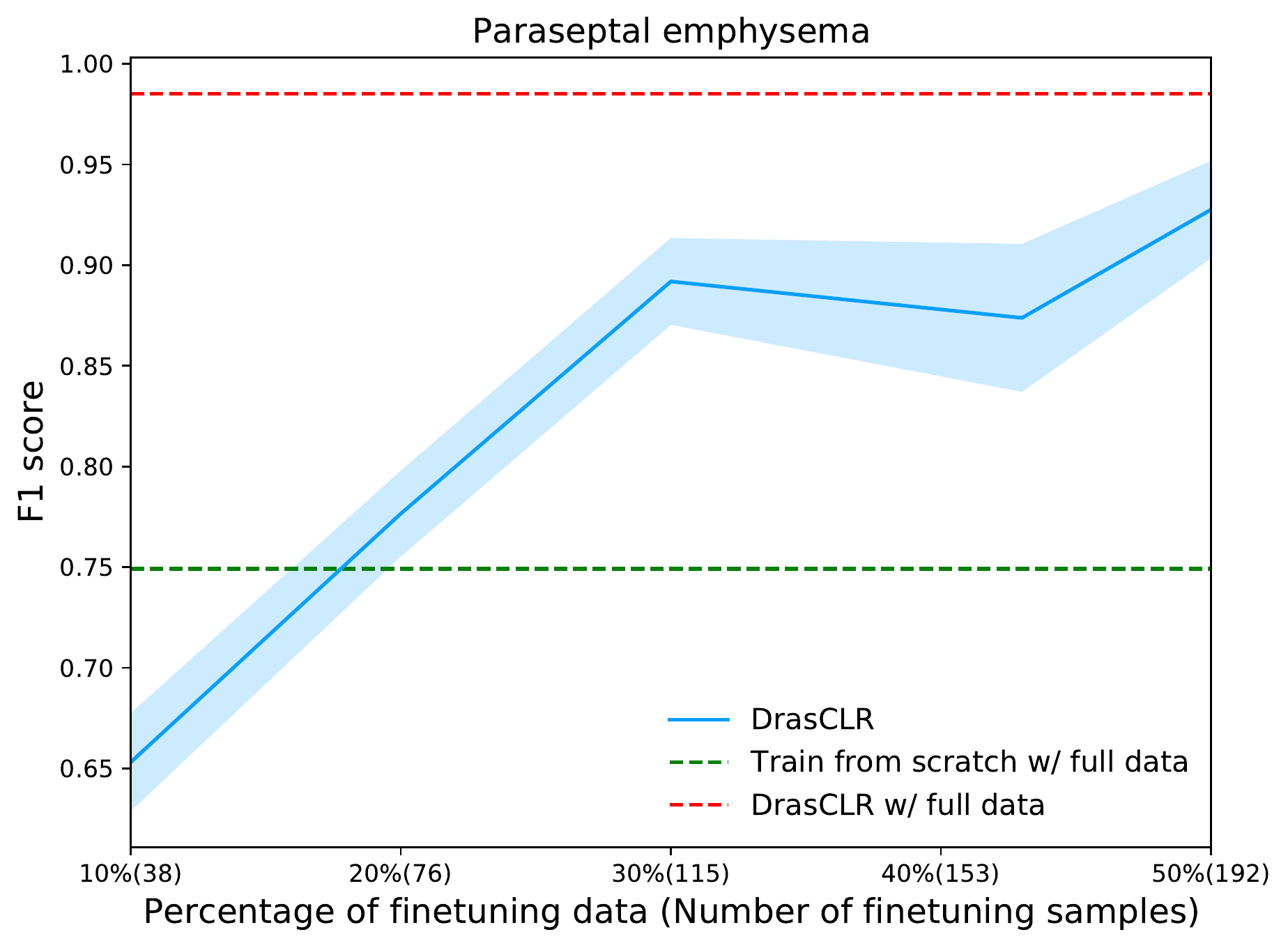}
\end{subfigure}
    \caption{Results of fine-tuning with different amounts of data. We perform evaluations for centrilobular (left) and paraseptal (right) emphysema detection. Compared to the model fine-tuned with full data from scratch (random initialization), the DrasCLR pre-trained model only needs $50\%$ and $20\%$ annotated data to achieve the same performance for centrilobular and paraseptal emphysema detection, respectively.
    }
    \label{fig:error_band}
\end{figure*}

Transfer learning makes use of the knowledge of underlying data structure learned by the pre-trained models and has been demonstrated to be beneficial in medical imaging analysis, where the amount of annotated data is often limited. We simulate the scenarios of using a subset of annotated data to investigate the power of our method in transfer learning. Specifically, we fine-tune the DrasCLR pre-trained model by starting with $10\%$ annotated emphysema patches and gradually increasing the amount of annotations by $10\%$ in subsequent experiments. Fig.~\ref{fig:error_band} shows the results of transfer learning on two target tasks. The performance of centrilobular detection learning from scratch with the entire dataset can be surpassed using DrasCLR with only $50\%$ of the dataset, hence doubling the annotation efficiency. The performance of paraseptal detection learning from scratch with the entire dataset can be surpassed using DrasCLR with only $20\%$ of the dataset, thus improving the annotation efficiency by five times. These results demonstrate how DrasCLR can significantly reduce the cost of manual image annotation, ultimately leading to more label-efficient deep learning.

\subsection{Ablation Study}
In this section, we conduct ablation experiments to validate the effects of several DrasCLR components.

\subsubsection{Design Choices for Incorporating Anatomical Location}
We demonstrate the effects of various designs for incorporating anatomical location information into DrasCLR. Specifically, we compare the following four approaches:
(1) No conditioning, which uses a standard CNN to extract features from image patches without taking their anatomical locations into account;
(2) Concatenation~\cite{sun2021context}, which concatenates the features from the last layer of the standard CNN with patch location and then fuses them through fully connected layers;
(3) HyperNetwork~\cite{ha2016hypernetworks}, which uses a separate neural network that takes an anatomical location as input to generate weights in the standard CNN;
(4) A CNN consists of our proposed Loc-CondConv layers. All models in this ablation experiment are trained using only local contrastive loss. As shown in Table.~\ref{tbl:ablation_condition}, the simple concatenation approach outperforms the standard CNN in three image-level tasks, suggesting that including anatomical location enriches the learned representations and enhances the performance of the downstream analysis. Furthermore, Table.~\ref{tbl:ablation_condition} shows that the CNN with Loc-CondConv layers achieves the best performances in all target tasks, demonstrating it is a superior design for incorporating anatomical location information.
 
\begin{table*}[t]
\caption{Ablation study for how to incorporate anatomical location. We report R-Square for $\log$\texttt{FEV1pp}, accuracy scores for the GOLD and CLE scores. The mean and standard deviation values are calculated via 5-fold cross-validation.}
\vspace{5px}
\centering
\begin{tabular}{lccc}
\toprule
Method&$\log$\texttt{FEV1pp}&GOLD&CLE\\
\midrule
No conditioning&$0.57_{\pm.04}$&$61.8_{\pm1.1}$&$48.0_{\pm.9}$\\
Concatenation&$0.60_{\pm.01}$&$62.5_{\pm1.0}$&$49.2_{\pm1.1}$\\
HyperNetwork&$0.60_{\pm.01}$&$58.6_{\pm1.7}$&$44.1_{\pm1.3}$\\
\midrule
Loc-CondConv (Ours)&\bm{$0.62_{\pm.02}$}&\bm{$63.4_{\pm1.0}$}&\bm{$50.3_{\pm.9}$}\\
\bottomrule
\end{tabular}
\label{tbl:ablation_condition}
\end{table*}
 
\subsubsection{Impact of Anatomical Location for Disease Detection}
We have thus far validated the importance of anatomical location as well as the effectiveness of Loc-CondConv layer for image-level prediction tasks. To investigate whether anatomical location is an important factor in representing fine-grained local features, we explore the impact of perturbing the input locations of DrasCLR on the performance of emphysema detection. In particular, instead of extracting features from a given patch using its correct anatomical location as input, we extract features from the patch using a random location in the lung. We employ a linear readout fine-tuning scheme on the same set of patches as described in Section~\ref{sec: emphysema_detection}.
As shown in Table~\ref{tbl:ablation_sensitivity}, by using random anatomical locations as inputs, the detection accuracy drops by $6\%$ and $16\%$ for centrilobular and paraseptal emphysema,  respectively, showing statistically significant decreases (p-value$<0.005$, one-sided two sample t-test).
The results indicate that the DrasCLR pre-trained encoder is sensitive to anatomical locations and captures anatomy-specific features. 

\begin{table}[t]
 \caption{Sensitivity of the pre-trained DrasCLR model to anatomical location perturbation. The results are the means and standard deviations of emphysema detection accuracy for 5-fold cross validation.}
 \vspace{10px}
 \centering
  \begin{tabular}{lcc}
  \toprule
   Subtype&Random location&Patch location\\
   \midrule
Centrilobular emphysema&$73.3_{\pm2.6}$&\bm{$79.2_{\pm1.4}$}\\
Paraseptal emphysema&$56.4_{\pm4.1}$&\bm{$72.1_{\pm4.3}$}\\
  \bottomrule
  \end{tabular}
  \label{tbl:ablation_sensitivity}
 \end{table}

\subsubsection{Neighboring Contrastive Loss}
\label{sec:ablation_neighbors}
In this experiment, we investigate the effect of neighboring contrastive loss as well as the impact of the number of neighbors used. 
We pre-train DrasCLR models with no neighboring contrast as well as with different numbers of neighbors. Using a linear readout scheme, we benchmark the performance of the pre-trained models in downstream image-level tasks. As shown in Table~\ref{tbl:ablation_condition_num}, the incorporation of spatial context from neighboring patches enhances the performance of image-level tasks in most situations. The improvement is particularly notable in the prediction of the centrilobular visual score, which ranges from $5\%$ to $8\%$ depending on the number of neighbors. This is likely due to the fact that CLE grades are determined by the extent to which the lung's center is damaged by the disease, and the neighboring contrasting strategy encourages the learning of common disease patterns that span multiple anatomical regions.
 
 \begin{table}[t]
 \caption{ Ablation study for neighboring contrastive loss. We report R-Square for $\log$\texttt{FEV1pp}, accuracy scores for GOLD and CLE scores. The mean and standard deviation values are calculated via 5-fold cross-validation.}
 \vspace{10px}
 \centering
  \begin{tabular}{lccc}
  \toprule
   Method&$\log$\texttt{FEV1pp}&$\log$\texttt{FEV1pp/FVC}&CLE\\
   \midrule
No Neighboring Contrast&$0.62_{\pm.02}$&$0.69_{\pm.02}$&$50.3_{\pm0.9}$\\
\# Neighbors = 1&$0.61_{\pm.01}$&$0.70_{\pm.01}$&$53.0_{\pm1.3}$\\
\# Neighbors = 2 (Ours)&\bm{$0.63_{\pm.01}$}&\bm{$0.71_{\pm.01}$}&\bm{$53.9_{\pm0.8}$}\\
\# Neighbors = 3&\bm{$0.63_{\pm.01}$}&\bm{$0.71_{\pm.01}$}&$52.5_{\pm0.5}$\\
  \bottomrule
  \end{tabular}
  \label{tbl:ablation_condition_num}
 \end{table}

\section{Discussions}
\label{sec:discussions}
\subsection{Does DrasCLR learn disease-related features?}
In this paper, we propose a novel self-supervised method, DrasCLR to learn \emph{disease-related} representation of 3D medical images. Medical images have recurring and similar anatomy across patients. While previous self-supervised methods~\cite{zhou2021models, haghighi2021transferable} used this knowledge to learn common anatomical representation, our method uses this knowledge to create hard negative samples  in contrastive learning. As a result, our method is more sensitive to tissue abnormalities and can encode more disease-related information. As seen in Table~\ref{tbl:COPD}, our method demonstrates excellent performance in predicting a wide range of clinical variables, such as spirometry measures and COPD phenotypes, which are closely associated with the degree of lung impairments in COPD patients. Emphysema is a hallmark of COPD. However, visually assessing emphysema at CT is time-consuming and subject to human variability. Our method demonstrates superior performance for this task and may provide a data-driven way of quantifying the visual score of emphysema from CT imaging. 

Moreover, as evidenced by Table~\ref{tbl:concordance}, imaging features learned by our method add incremental value to the BODE index for survival analysis in COPD patients, suggesting that our method is capable of capturing complementary risk factors from CT imaging. In addition to the evaluation on the COPD cohort, we demonstrate, as shown in Table~\ref{tbl:RU}, that DrasCLR can learn robust features associated with COVID-19 severity. The ability to represent disease-related information from medical images in an unsupervised manner is particularly useful during pandemic outbreaks, when labeled data is rarely available. 

\subsection{Does DrasCLR extract location-specific features?}
A conventional convolutional layer is designed to be transitionally invariant. However, pathological patterns tend to be heterogeneous across locations in the human body, and an \emph{one-size-fits-all} design may not be sufficient to learn the variety of tissue abnormalities at different anatomical locations. In this paper, we incorporate anatomical context into representation learning via two components in DrasCLR. First, image registration is used to provide a unified anatomical coordinate system, and all images are aligned to the atlas. Second, a novel Loc-CondConv layer is introduced to have modifiable weights that are conditionally dependent on anatomical location.

We have explored different ways to incorporate anatomical location into representation learning. As seen in Table~\ref{tbl:ablation_condition}, the Loc-CondConv layer is superior to simple concatenation and hypernetwork. To further validate the effect of our design, we investigate the impact of perturbing the input locations of DrasCLR on emphysema detection. As shown in Table.~\ref{tbl:ablation_sensitivity}, the emphysema detection accuracy decreases when random location is used, demonstrating that lack of correct anatomical context is detrimental. 

\subsection{Why do we choose sliding-window based approach for emphysema segmentation?}
The most well-known architecture for medical image segmentation is U-Net~\cite{ronneberger2015u}, which is composed of opposing convolution and deconvolution layers, and spatial information is provided through skip connections to each decoder layer to recover fine-grained details. However, training U-Net requires the data to have complete pixel/voxel-level annotation, which is difficult to obtain. The emphysema annotation in the COPDGene dataset was collected as bounding boxes by a physician clicking on regions surrounding by pathological tissues, thus the annotation is not complete at the voxel level. Our method can leverage these partially annotated images and produce voxel-wise emphysema classification in a sliding-window fashion. Moreover, the DrasCLR pre-trained model is capable of utilizing spatial information by modifying kernels' weights based on the input anatomical location. As illustrated by Fig.~\ref{fig:emph_volume}, our method can produce high-quality emphysema segmentation in which the regional distribution of detected emphysema matches with the clinical description of emphysema subtypes. The box plots in Fig.~\ref{fig:volume_proportion} show quantitative evidence that the detected emphysema volume correlates with the subjects' COPD stages.

\section{Conclusion}
\label{sec:conclusion}
In this paper, we present a novel method for anatomy-specific self-supervised representation learning on 3D medical images. We propose two domain-specific contrasting strategies to learn disease-related representations, including a local contrasting loss to capture small disease patterns and a neighboring contrasting loss to learn anomalies spanning across larger anatomical regions. 
In addition, we introduce a novel conditional encoder for location-specific feature extraction. The experiments on multiple datasets demonstrate that our proposed method is effective, generalizable, and can be used to improve annotation efficiency for supervised learning.

\section*{Acknowledgments}
This work was partially supported by NIH Award Number
1R01HL141813-01 and NSF 1839332 Tripod+X. We are grateful for the computational
resources provided by Pittsburgh SuperComputing grant number TG-ASC170024.

\bibliographystyle{unsrt}  
\bibliography{references}

\end{document}